%% file: acl2020.tex
\pdfoutput=1 
\documentclass[11pt,a4paper]{article}
\usepackage{authblk}
\usepackage{subcaption}
\usepackage[hyperref]{acl2020}
\usepackage{graphicx}
\usepackage{times}
\usepackage{url}
\usepackage{hyperref}
\usepackage{longtable}
\usepackage{array,multirow,graphicx}
\usepackage[thinlines]{easytable}
\usepackage{geometry}

\usepackage{microtype}
\usepackage{wrapfig}
\usepackage{longtable}
\usepackage{float}
\usepackage{supertabular}
\usepackage{afterpage}

\aclfinalcopy

\title{Roboflow 100: A Rich, Multi-Domain Object Detection Benchmark}

\author[*]{Floriana Ciaglia}
\author[*]{Francesco Saverio Zuppichini}
\author[*]{Paul Guerrie}
\author[*]{Mark McQuade}
\author[*]{Jacob Solawetz}
\affil[*]{{\fontfamily{qcr}\selectfont
[floriana,paul,francesco,mark,jacob]@roboflow.com
}}

\date{}

\begin{document}
\maketitle

\begin{abstract}
    The evaluation of object detection models is usually performed by optimizing a single metric, e.g. mAP, on a fixed set of datasets, e.g. Microsoft COCO and Pascal VOC. Due to image retrieval and annotation costs, these datasets consist largely of images found on the web and do not represent many real-life domains that are being modelled in practice, e.g. satellite, microscopic and gaming, making it difficult to assert the degree of generalization learned by the model.

    We introduce the Roboflow-100 (RF100) consisting of 100 datasets, 7 imagery domains, 224,714 images, and 805 class labels with over 11,170 labelling hours. We derived RF100 from over 90,000 public datasets, 60 million public images that are actively being assembled and labelled by computer vision practitioners in the open on the web application \href{https://universe.roboflow.com/}{Roboflow Universe}. By releasing RF100, we aim to provide a semantically diverse, multi-domain benchmark of datasets to help researchers test their model's generalizability with real-life data. RF100 download and benchmark replication are available on \href{https://github.com/roboflow-ai/roboflow-100-benchmark}{GitHub}.

\end{abstract}

\input{tables/datasets-table-grouped}

\section{Introduction}
In object detection research, Microsoft COCO\cite{coco} and Pascal VOC\cite{pascalVOC} have become the \textit{de facto} benchmark standards to train and evaluate the performance of models. These models are then released to the public who fine-tuned them on a smaller dataset consisting of specific imagery and targets of interest. While the general object detection benchmarks provide a proxy for how the model will perform in a similar setting, there is no substitute for domain-specific training. 

There is a strong research interest in evaluating models on a more diverse task set. For example, 13 Roboflow community open source datasets have organically been used by researchers to create The Object Detection in the Wild (ODINW) \cite{glip} benchmark. This benchmark was utilized to assert model performance for object detection and zero-shot capabilities in Florence \cite{florence} and GLIP \cite{glip}. Through curating a larger set of narrow task datasets, we build on this interest for deeper domain-specific assessment and introduce a stronger benchmark we called Roboflow 100 (RF100). 

RF100 consists of a collection of 100 crowd-sourced object detection (OD) datasets, specifically constructed by Roboflow users to represent a chosen subject of study. In Figure \ref{fig:dataset-examples} we show a small sample of annotated image domains that the RF100 benchmark encompasses, which range from satellite and aerial assessment to microscopic image analysis.

By introducing a benchmark of narrow subject matter datasets, we accomplish two goals. Firstly, our benchmark provides a strong collection of closed-domain tasks used in the wild that are of demonstrated interest to practitioners for researchers that are designing models with the intent of them being fine-tuned. Secondly, researchers building general models can test transfer between object detection tasks in zero-shot or few-shot fashion. 

\input{figures/figure-samples}

\section{Related Work}

Historically, object detection datasets are created by gathering a large corpus of images and sourcing annotators to label objects in a fixed set of classes.

The Pascal VOC project \cite{pascalVOC} is a collection of datasets made to enhance object detection tasks and encourage researchers to create models that recognize objects in realistic scenes in the form of a challenge. The Pascal VOC challenges started in 2005 and laid the groundwork for a new generation of state-of-the-art benchmarks.

%ImageNet - large old school, for classification
ImageNet \cite{deng2009imagenet} is an image dataset comprised of over 14 million images each described by word phrases called "synset". It was specifically created to answer the need in the industry for a high-quality object categorization benchmark with clearly established evaluation metrics. Similarly to RF100, ImageNet was created to encourage the creation of more generalizable machine learning models.

%Open Images - made to span semantics across categories for pretraining - but like other pretraining datasets, it is labelled on mostly web images, not from people who are making industrial applications from their roboflow projects.
Open Images \cite{OpenImages} is an image collection with over 9 million annotated images and 600 object classes. It was created to enable the study of tasks such as image classification, object detection, visual relationship detection, instance segmentation, and multi-modal image descriptions all from one joined resource to stimulate progress towards image scene comprehension.

The Common Objects in Context \cite{coco} (COCO) benchmark is a large-scale object detection and segmentation dataset with a total of 2.5M labelled instances in 328K images. Ever since its release, the COCO benchmark has established itself as the state-of-the-art standard thanks to its abundance of subjects and annotations.

The Objects365 \cite{objects365} dataset is a large collection of separate object detection datasets that is comprised of images from the website Flicker. Images are gathered, categories are assigned, and then a labelling team is employed to create annotations. Pretraining models on Objects365 are shown to benefit performance on downstream tasks, such as COCO.

The Object Detection in the Wild (ODINW) \cite{glip} dataset was released in the same spirit as RF100. The original ODINW version uses 13 Roboflow object detection datasets to assess the generalizability of their zero-shot model, GLIP \cite{glip}. Despite advances in open vocabulary object detectors like GLIP, accurate and fast object detection still requires custom training to be performed on quality, annotated data with a closed vocabulary.

Unlike prior related work, we assemble a large-scale object detection dataset that is sourced via image upload and annotation by practitioners on a web application who are using computer vision to accomplish real tasks.

\section{Methods}
In this section, we describe the RF100 dataset creation process and our initial modelling experiments on the new benchmark.

\subsection{Dataset Collection and Distribution}

\href{https://universe.roboflow.com/}{Roboflow Universe} is a public repository of computer vision dataset that over 100,000 Roboflow users have assembled and labeled for their own custom use cases. We selected 100 datasets from \href{https://universe.roboflow.com/}{Roboflow Universe} for our benchmark using the following criteria:

\begin{itemize}
    \item Effort - the user spent substantial labelling hours working on the task;
    \item Diversity - the user was working on a novel task;
    \item Quality - the user annotated with high fidelity to the task;
    \item Substance - the user assembled a substantial dataset with nuance;
    \item Feasibility - the user was attempting a learnable task;
\end{itemize}

After selection for inclusion, all datasets were processed in the following way:
\begin{itemize}
    \item All images resized to 640x640 pixels following best practices \cite{yolov7}
    \item Eliminate class ambiguity, i.e. if a class was labeled by the original author as \textit {0} to represent a flower, that class label would be changed to a word descriptive of the actual subject such as \textit {daisy}
    \item The train, validation, and test split were manipulated only in the instances where either one or more split sets were missing completely, or where one, or more, of the split sets were extremely under-represented. In all other cases, we respected the split ratio chosen by the original author of the dataset.
    \item Underrepresented classes were filtered when they represented less than 0.5 percent of all objects in a dataset. These classes tended to be labeling errors.
\end{itemize}
The datasets are available for download from \href{https://github.com/roboflow-ai/roboflow-100-benchmark}{GitHub} or from the \href{https://universe.roboflow.com/roboflow-100}{Roboflow Universe website} by clicking on the \textit{Export} button.

\input{tables/stats-datasets-grouped}

\subsection{Semantics}
We selected seven different semantic categories to achieve comprehensive coverage of real-life possible domains: Aerial, Video Games, Microscopic, Underwater, Documents, Electromagnetic and Real World. 

The Real World category is the biggest in the RF100 benchmark since the majority of use cases for computer vision involve everyday scenes. We included indoor, outdoor, Vehicles, animals, plants, damage control, safety, electronics, geology, board games and various human activity images.

The Video Games category includes virtual reality scenes, robot fighting, first-person shooters and MOBA; The Underwater category includes fishery sights from both seas and aquariums, as well as inanimate objects found underwater (i.e. pipes).

The Microscopic category is comprised of items that can only be seen with the aid of a microscopic lens like bacteria and human cells; the Aerial category includes images taken from an overhead view, including images from space and drones; the Electromagnetic category includes scenarios where electromagnetic waves were used to capture the pictures,  X-rays, MRIs, IR, thermal and night vision cameras; and finally, in the Documents category, we include all images that relate to articles, papers, tables, diagrams and social media.

Figure \ref{fig:categories} shows samples for each category.

\subsection{Data Statistics and Analysis}
Table \ref{table:stats-datasets-grouped} summarizes the benchmark's metadata including the number of datasets, images and classes present in each category. Per dataset statistics and results can be seen in Table \ref{table:datasets-stats}.

Figure \ref{fig:dataset-stats} reports different statistics grouped by category, such as number of classes and bounding boxes area. Most notably,  \textit{Aerial}, \textit{Microscopic} and \textit{Electromagnetic} have smaller bounding boxes compared to the rest. Moreover, the average number of classes across different categories is only ten, meaning in practice people need to identify a small set of objects.

\input{figures/stats.tex}
\input{figures/scatter.tex}

Figure ~\ref{fig:scatter-plot} shows a scatter plot produced to analyze the vector clustering degree of the RF100 categories using CLIP embeddings \cite{clip} generates for each of its datasets. This illustration shows that the datasets in each category do tend to cluster together. You can also view an image-level clustering of these semantics on the \href{https://www.rf100.org/}{RF100 web exploration web demo}.

\subsection{Experiments and Evaluation}
% Construction and Analysis of the chosen experiments.  
We trained popular object detection model architectures on RF100 and report the results. Only one model instance was trained per dataset.
\input{tables/results-datasets-grouped} 

\paragraph{Finetuning:}

We finetuned two comparable versions of YOLOv5 \cite{glenn_jocher_2020_4154370} and YOLOv7\cite{yolov7}: YOLOv5s and YOLOv7, with 7.2M parameters and 36.9M parameters respectively and similar FPS when evaluated on a Tesla V100.

We trained both models with default hyperparameters for 100 epochs at 640x640 resolution.

\paragraph{Zero Shot:}
We also evaluated GLIP \cite{li2022grounded}, a zero-shot detector that can solve open vocabulary detection by rephrasing it as a grounding task, in which each noun in a sentence is associated with an object in an image.

GLIP's default behaviour for the evaluation task is to use the objects' class names as the evaluation prompts. We recognize that some of the RF100 datasets might not have very descriptive class names and that our imagery is, as intended, more specific than other more general benchmarks. As a consequence, we believe that providing GLIP with word phrases to describe the objects in our images would be a more fair challenge for a zero-shot model. We compiled a list of descriptions that match each class for most of the datasets in Roboflow-100. The datasets we did not include in the prompt remapping already had sensible class names.

We evaluate RF100 on the GLIP-T model pre-trained on O365, GoldG, Conceptual Captions 3M, and SBU captions \cite{li2022grounded}.

Results are reported in Table  \ref{table:results-datasets-grouped}.

\section{Discussion}

Our initial benchmarks show that there is variation in model performance between models across datasets and datasets domains that may run contrary to the model's ranking on incumbent benchmarks. A given model may perform better on one dataset and worse on another. While we do not investigate the underlying reasons for performance differentials, our results suggest that there are likely significant improvements to be made to object detection models to expect a wider array of custom datasets that they may be applied to.

Lastly, our evaluation shows that zero-shot object detection models lose considerable performance when extended to new domains. While evaluation on COCO shows that zero-shot models like GLIP are approaching the performance of their fine-tuned counterparts, our benchmark shows that they do not generalize well to new, more obscure imagery domains. This is likely due to the out-of-vocabulary imagery types and class titles from what frequently appears in image datasets assembled on the web.

\section{Conclusion}

We introduce the RF100 object detection benchmark of 100 datasets to encourage the evaluation of object detection model performance to test model generalizability across a wider array of imagery domains. Our initial evaluation shows that the new RF100 benchmark will provide valuable insights into how new object detection models will perform in the wild. RF100 is available for download on \href{https://github.com/roboflow-ai/roboflow-100-benchmark}{GitHub}.

\section*{Acknowledgments}
We thank all of the advisors we have had on our research both internally at Roboflow, and externally in the machine learning community. We would also like to thank Intel for sponsoring the work involved in constructing the RF100 benchmark. 

Finally, we thank everyone working on public computer vision datasets on Roboflow Universe and in particular, the creators of the RF100 datasets: Abhishek Dada, Adam Crenshaw, Adrian Rodriguez, Ahmad Rabiee, Alex Hyams, Aman Ahuja and Alan Devera, Ammar Abdlmutalib, Amro, Anshul Rankawat, Kapil Verma, Shubhankar Rawat, Manisa Mondal, Pranav Arora, Arfiani Nur Sayidah, Brad Dwyer, Brad Dwyer, CC Moon, Chang Yuan, Dane Sprsiter, David Lee, Djamel Mekhlouf, Abrisse Cerine, Anfal Lanna, Yasmin Emekhlouf, Evan Kim, MJ Kim, Graham Doerksen, Ilyes Talbi, Jan Douwe, Jason Zhang, Caden Li, Jhonathann, Joao Paulo Martins, Jordan Bird, Leah Bird, Carrie Ijichi, Aurelie Jolivald, Salisu Wada, Kay Owa, Chloe Barnes, Joseph Nelson, Brad Dwyer, and Cheng Hsun Teng , Justin Henke, Reginald Viray, Kais Al Hajjih, Karen Weiss, Kat Laura, Lao and Shiguang, Lukas D. Ringle, Matteo Pacini, Melanie S. Capalungan, B-Jay Daguio, Isaac Balbuena, and Reanne Joy Rafael, Mevil Crasta, Miguel Fernández Cruchaga, Mike Drickramer, Minoj Selvaraj, Mohamed Attia, Mohamed Refai, Abarna, Amjad Hafiz, Sutheshan Maiu, and Thanusha Sritharan, Mohamed Sabek, Monika Patel, Kartik Attri, Aniket Dhanotia, Divyam Jha, Pankaj, kanchan, Ujjwal Sharma, Garvita Vijay, Aniket Choudhary, Pearl Rathour, Roshni Ghai, Kavya Shukla, Preeti Sharma, Ananya Kharayat, Krishna Gambhir, Lav Naruka, Kas, Sejal, Tejasvi Singh, Ayush Sahu, Pri, Aniket Dhanotia, and Devansh Shrivastava , Montso Mokake, Muntaser Al Abdulla Aljouma, Nazmuj Shakib Diip, Afraim, Shiam Prodhan, NhiNguyen and Duong Duc Cuong , Nikita Manolis, Nirav Golyalla, Ali Fakhry, Ammar F, Shangyu, Nirmani, Oliver Giesecke, Christian Green, Omar Kapur,  Ricardo Jenez, Justin Jeng, and Jeffrey Day, Pablo Ochoa, Antonio Luna, Eliezer Alvarez, Paper Authors, Parfait Ahouanto, Pavel Kulikov, Djopa Volosata, Daria Podryadova, Pierrick Dossin, Ploylada Pharikarn, Potae, Phuthanig Areesawangkit, Prakasit, Tanchanok Prasootseangjan, Quandong Qian, Raya Al, Riccardo Secoli, Kaspars Sudars, Janis Jasko, Ivars Namatevs, Liva Ozola, and Niks Badaukis, Richard, Rik Biswas, Aakansha Prasad, Sarmistha Das, Rinat Landman, Ritesh Kanjee, Roopa Shree, Shriya J, Ruud Krinkels, Seokjin Ko, Simeon Marlokov, Sina, St Hedgehog Yusupov, T, Pankaj, Aniket Dhanotia, Tejasvi Singh, Garvita Vijay, Goku, Kavya Shukla, Aniket Choudhary, Ananya Kharayat, Monika Patel, Pearl Rathour, Devansh Shrivastava, Aniket Dhanotia, Priyansh Urajput, Sejal, Roshni Ghai, Krishna Gambhir, Ayush Sahu, Ujjwal Sharma, Divyam Jha, Kanchan, Kartik Attri, Lav Naruka, Kas, Preeti Sharma, Terada Shoma, Thuan Phat Nguyen, Vanitchaporn, Victor Perez, Stephen Groff, Mason Hintermeister, Wang Tianyi, Wilfred Shu and Adrian Stuart, Wojciech Blachowski, Wojciech Przydzial, Dorota Przydzial, Magdalena Przydzial-Mazur, and Bartlomiej Mazur, Xingwei He, Yilong Zheng, Yimin Chen, Yousef Ghanem, Yuanyu Anpei, Yudha Bhakti Nugraha and Kris, Yuntaewon, Hwanghyeyun,  Gimminseo, Gimnohyeon , Sindahong, Gimseongsu, Yuyang Li, Zhang Kaimin, Zhe Fan.

\pagebreak
\bibliographystyle{plainnat}
\bibliography{acl2020} 
\newpage
\appendix
\input{figures/categories.tex}
\newgeometry{left=1.7cm, top=1cm}
\input{tables/datasets-table}
\restoregeometry

\end{document}

%% file: tables/datasets-table-grouped.tex
\begin{table*}[]
    \setlength{\tabcolsep}{0.5em}
    {\renewcommand{\arraystretch}{1.2}
        \begin{center}
            \begin{tabular}{ c | c c c c c c c c}
                \hline
                \rule{0pt}{70pt}                          &
                \rotatebox[origin=l]{70}{Aerial}          & \rotatebox[origin=l]{70}{Videogames} & \rotatebox[origin=l]{70}{Microscopic} &
                \rotatebox[origin=l]{70}{Underwater}      &
                \rotatebox[origin=l]{70}{Documents}       &
                \rotatebox[origin=l]{70}{Electromagnetic} &
                \rotatebox[origin=l]{70}{Real World}      &
                \rotatebox[origin=l]{70}{COCO}
                \\ \hline
                Datasets                                  & 7                                    & 7                                     & 11    & 5     & 8     & 12    & 50     & 1      \\
                Images                                    & 9683                                 & 11579                                 & 13378 & 18003 & 24813 & 36381 & 110615 & 328000 \\
                Classes                                   & 24                                   & 88                                    & 28    & 39    & 90    & 41    & 495    & 80     \\ \hline
                YOLOv5                                    & 0.636                                & 0.859                                 & 0.650 & 0.560 & 0.716 & 0.689 & 0.752  & 0.568  \\
                YOLOv7                                    & 0.504                                & 0.796                                 & 0.591 & 0.662 & 0.722 & 0.607 & 0.699  & 0.697  \\ \hline
                GLIP                                      & 0.23                                 & 0.188                                 & 0.159 & 0.019 & 0.024 & 0.058 & 0.108  & 0.466  \\ \hline
            \end{tabular}
        \end{center}
        \caption{Overview of per-category metadata, including number of datasets, number of images, and number of classes across categories. Additionally, we record the average mAP@.50 value for the YOLOv5 and YOLOv7 models and the mAP@.50:.95 for the GLIP model for each category.}
        \label{table:datasets-stats-grouped}
    }
\end{table*}

%% file: figures/figure-samples.tex
\begin{figure*}[t]
	\centering
	\begin{tabular}{cccccc}
		\subcaptionbox{Aquarium \label{1}}{\includegraphics[width = 0.17\linewidth]{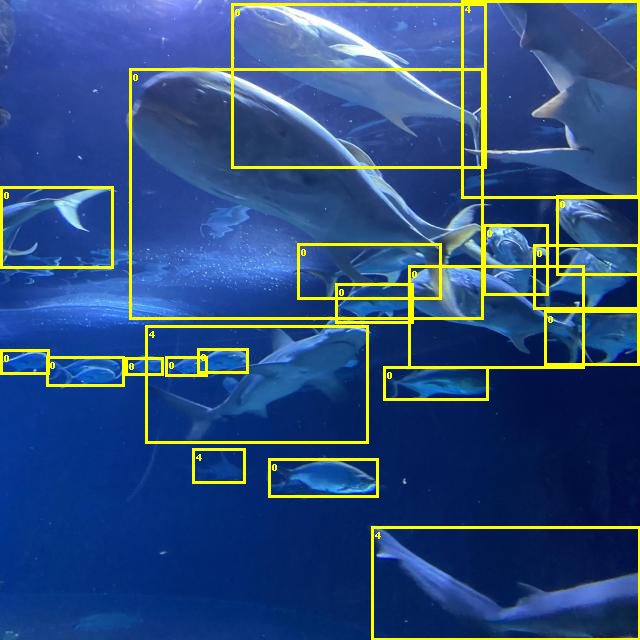}}                   &
		\subcaptionbox{Aerial \label{2}}{\includegraphics[width = 0.17\linewidth]{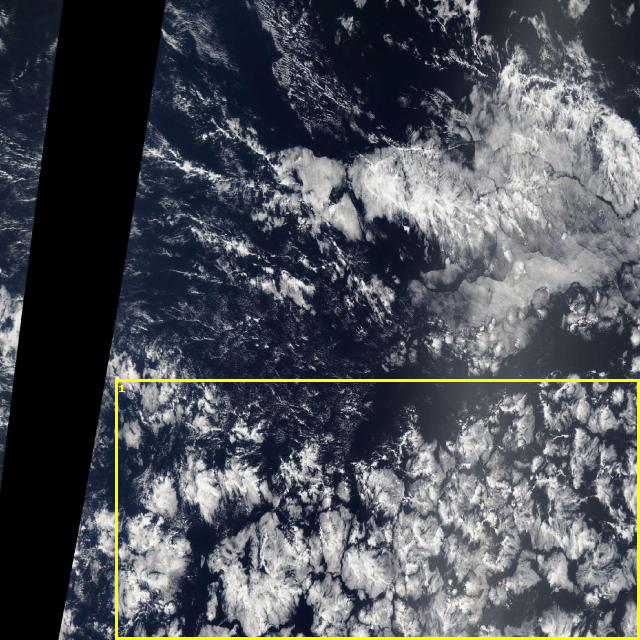}}              &
		\subcaptionbox{Bone Fracture \label{1}}{\includegraphics[width = 0.17\linewidth]{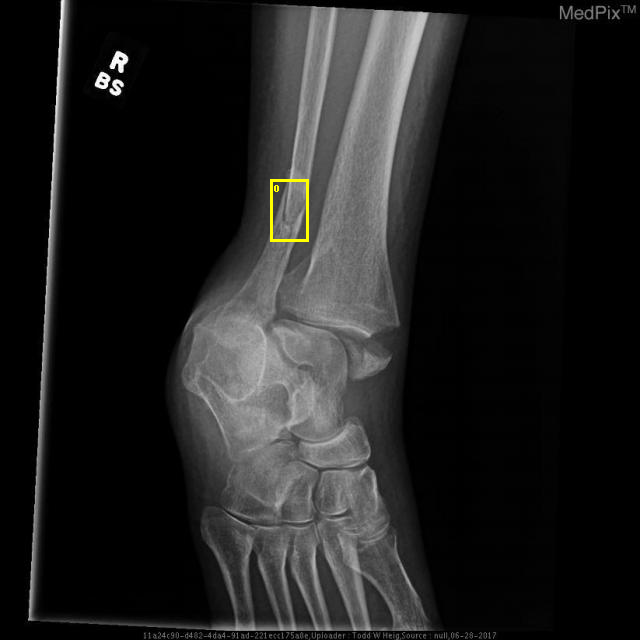}}          &
		\subcaptionbox{Brain MRI \label{2}}{\includegraphics[width = 0.17\linewidth]{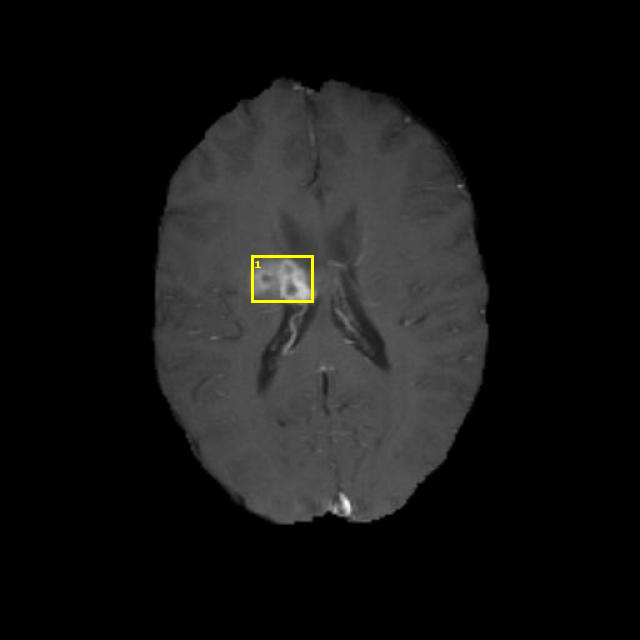}}                   &
		\subcaptionbox{Circuit Boards \label{2}}{\includegraphics[width = 0.17\linewidth]{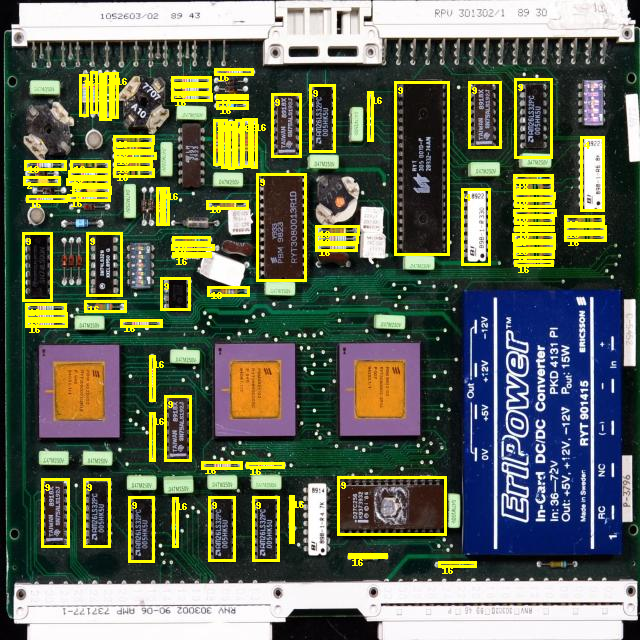}}         &
		\\
		\subcaptionbox{Crater Chains \label{2}}{\includegraphics[width = 0.17\linewidth]{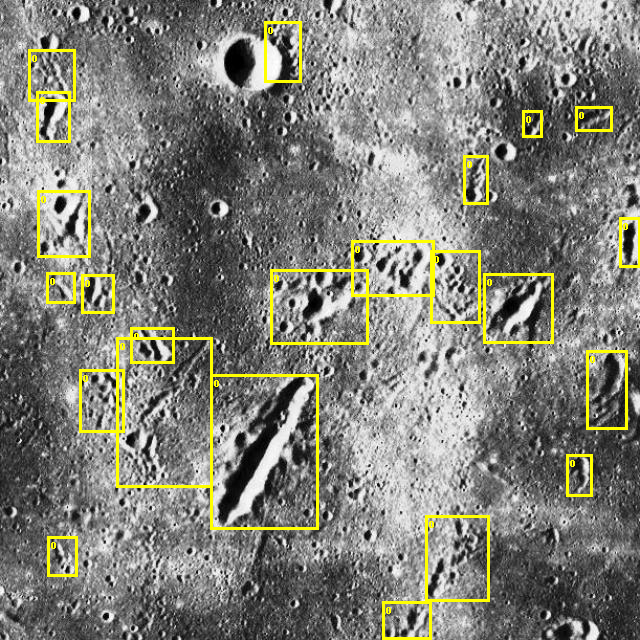}}           &
		\subcaptionbox{Liver Diseases \label{2}}{\includegraphics[width = 0.17\linewidth]{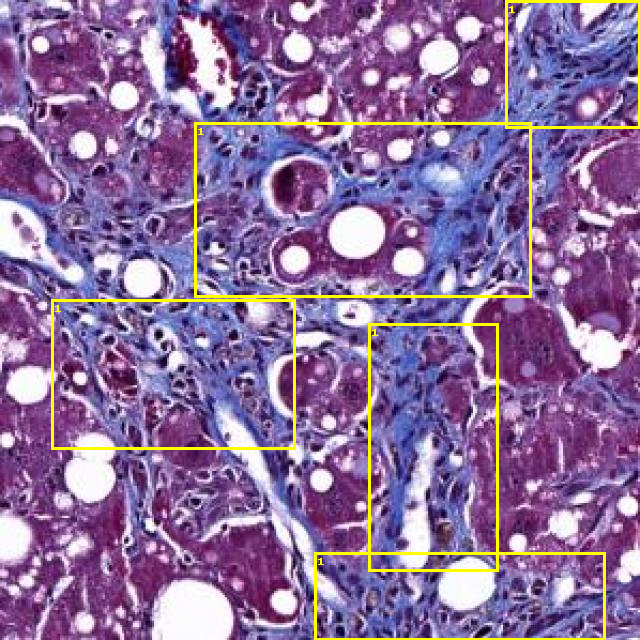}}          &
		\subcaptionbox{Microscopic \label{1}}{\includegraphics[width = 0.17\linewidth]{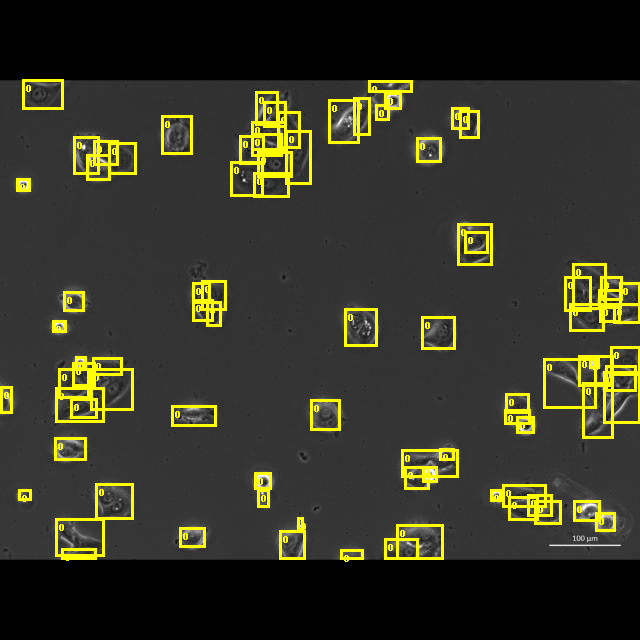}}              &
		\subcaptionbox{People in Painting \label{2}}{\includegraphics[width = 0.17\linewidth]{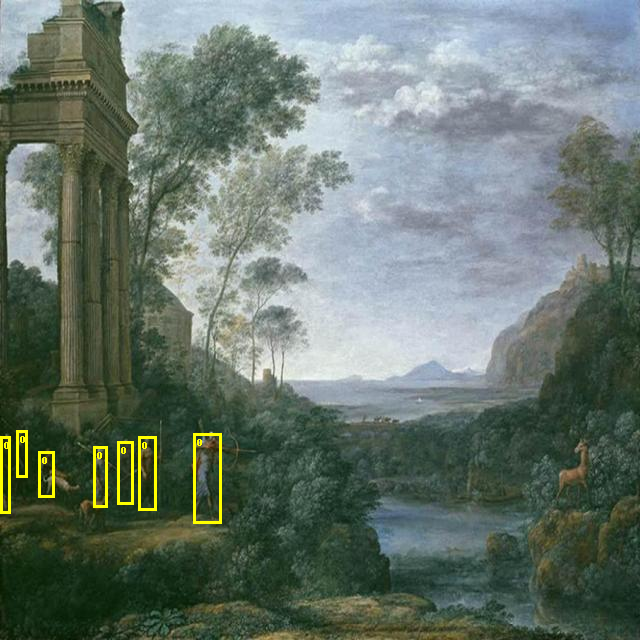}} &
		\subcaptionbox{Plant Diseases \label{1}}{\includegraphics[width = 0.17\linewidth]{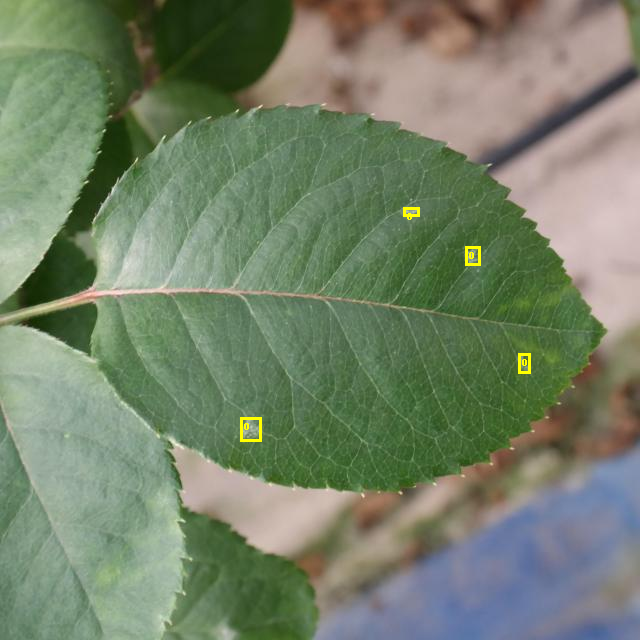}}          &
		\\
		\subcaptionbox{Robot Fighting\label{2}}{\includegraphics[width = 0.17\linewidth]{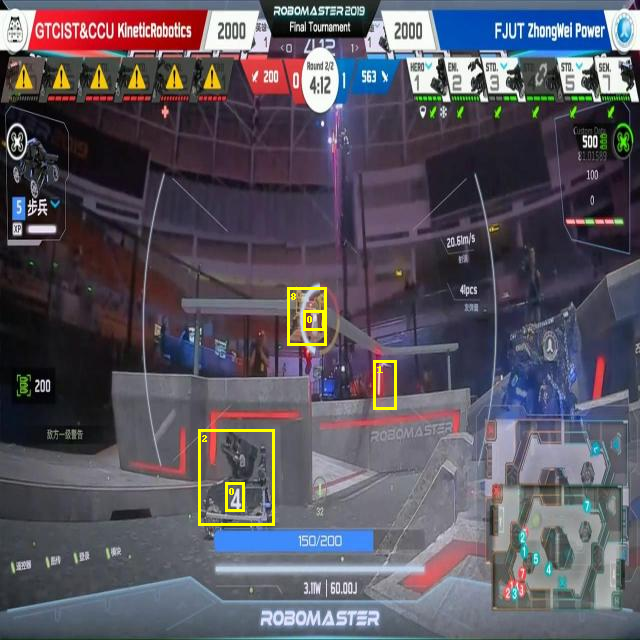}}          &
		\subcaptionbox{Thermal Dogs\label{2}}{\includegraphics[width = 0.17\linewidth]{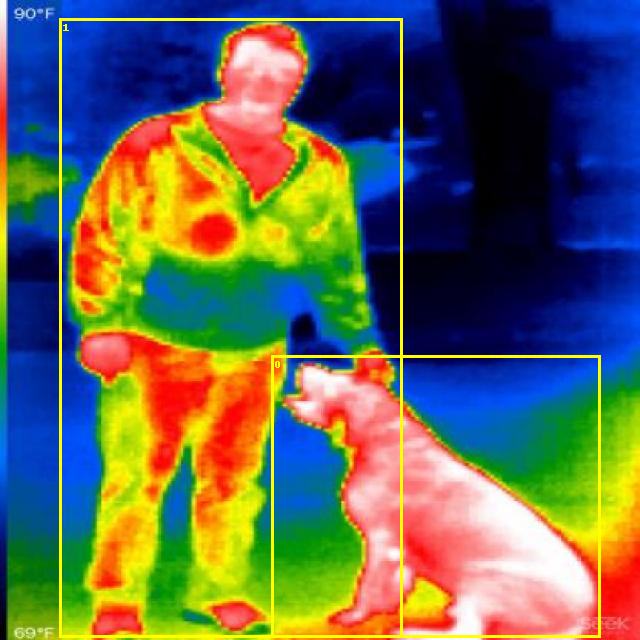}}              &
		\subcaptionbox{Tweets\label{2}}{\includegraphics[width = 0.17\linewidth]{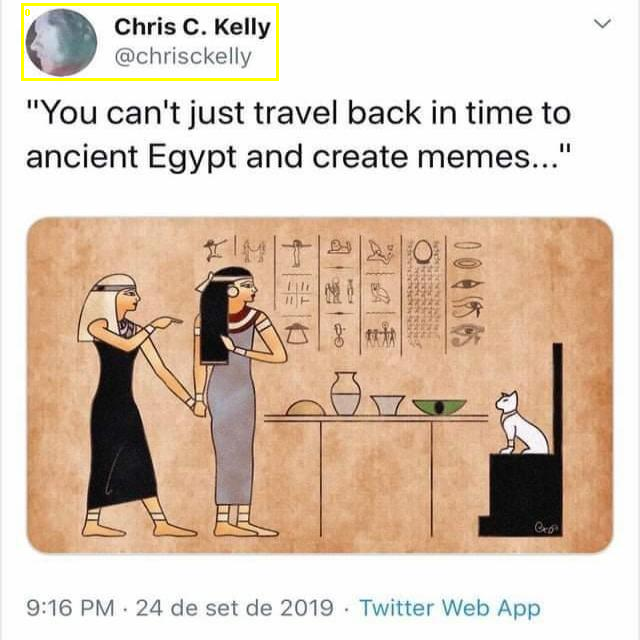}}                         &
		\subcaptionbox{Avatar Recognition\label{2}}{\includegraphics[width = 0.17\linewidth]{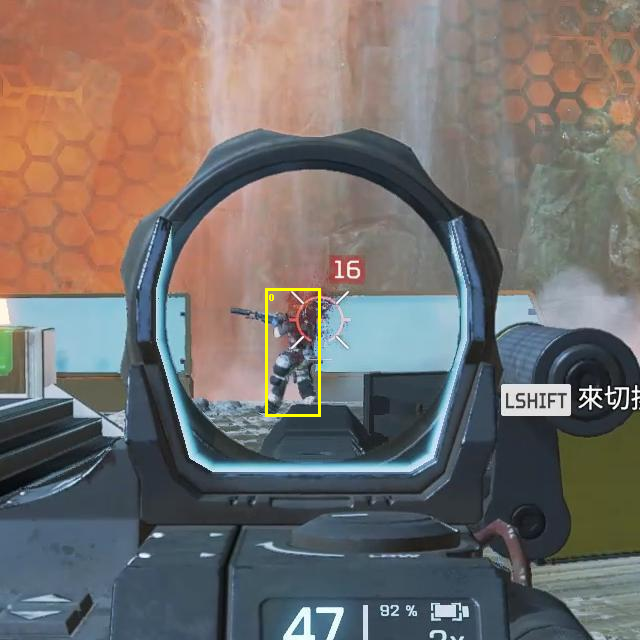}}  &
		\subcaptionbox{Night Cameras\label{2}}{\includegraphics[width = 0.17\linewidth]{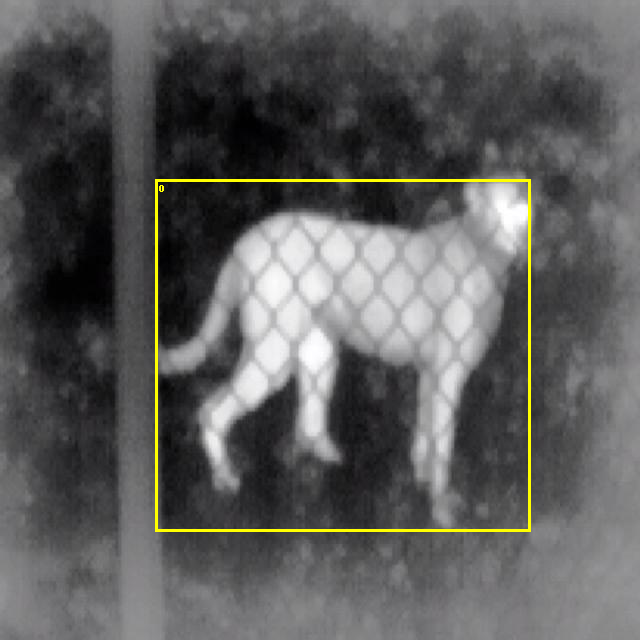}}            &
	\end{tabular}
	\caption{ Examples of annotated images in the RF100 benchmark. The data-set names are derived from the object of interested of each collection. Further examples can be found in Appendix \ref{appendix:A} \label{fig:dataset-examples} }
\end{figure*}

%% file: tables/stats-datasets-grouped.tex
\begin{table}[h]
    \begin{center}
    \setlength{\tabcolsep}{0.3em} % for the horizontal padding
        {\renewcommand{\arraystretch}{1.2}% for the vertical padding
            \begin{tabular}{ l | c c c}
                \hline
                Category        & Datasets   & Images         & Classes    \\
                \hline
                Aerial          & 7          & 9683           & 24         \\
                Videogames      & 7          & 11579          & 88         \\
                Microscopic     & 11         & 13378          & 28         \\
                Underwater      & 5          & 18003          & 39         \\
                Documents       & 8          & 24813          & 90         \\
                Electromagnetic & 12         & 36381          & 41         \\
                Real World      & 50         & 110615         & 495        \\
                \hline
                \textbf{Total}    & \textbf{100} & \textbf{224,714} & \textbf{805} \\
                \hline
            \end{tabular}
            \caption{Overview of per-category metadata, including number of datasets, number of images, and number of classes across categories.}
            \label{table:stats-datasets-grouped}
        }
    \end{center}
\end{table}

%% file: figures/stats.tex
\begin{figure}[H]
    \begin{center}
    \includegraphics[width=\linewidth]{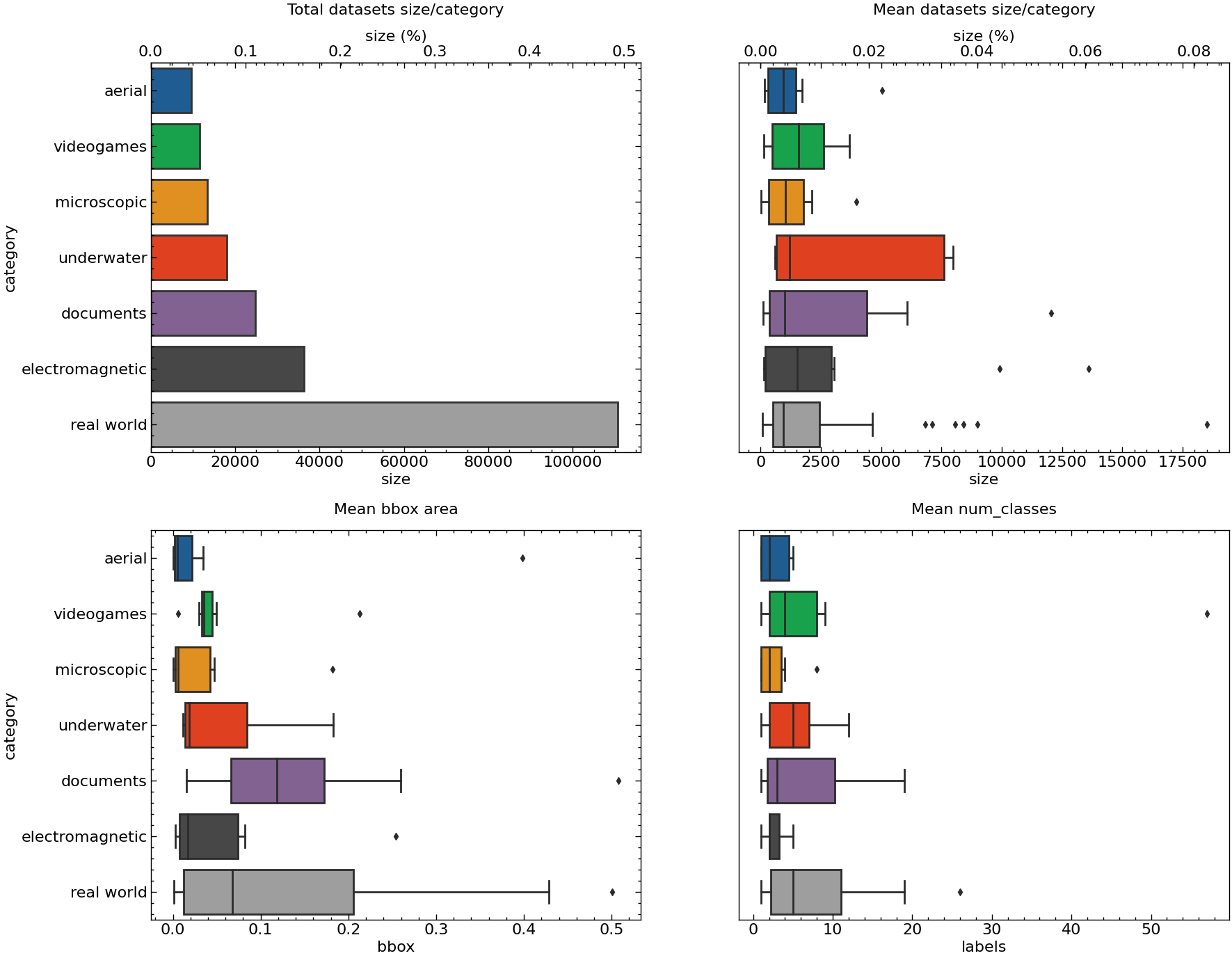}
    \caption{This figure shows different statistics for datasets grouped by category. Noticeably we can notice some categories (aerial, video games) have generally smaller bounding boxes compares to others, like documents. The number of classes is low, averaging 10.}
    \label{fig:dataset-stats}
\end{center}
\end{figure}

%% file: figures/scatter.tex
\begin{figure}[H]
    \centering\includegraphics[width=\linewidth]{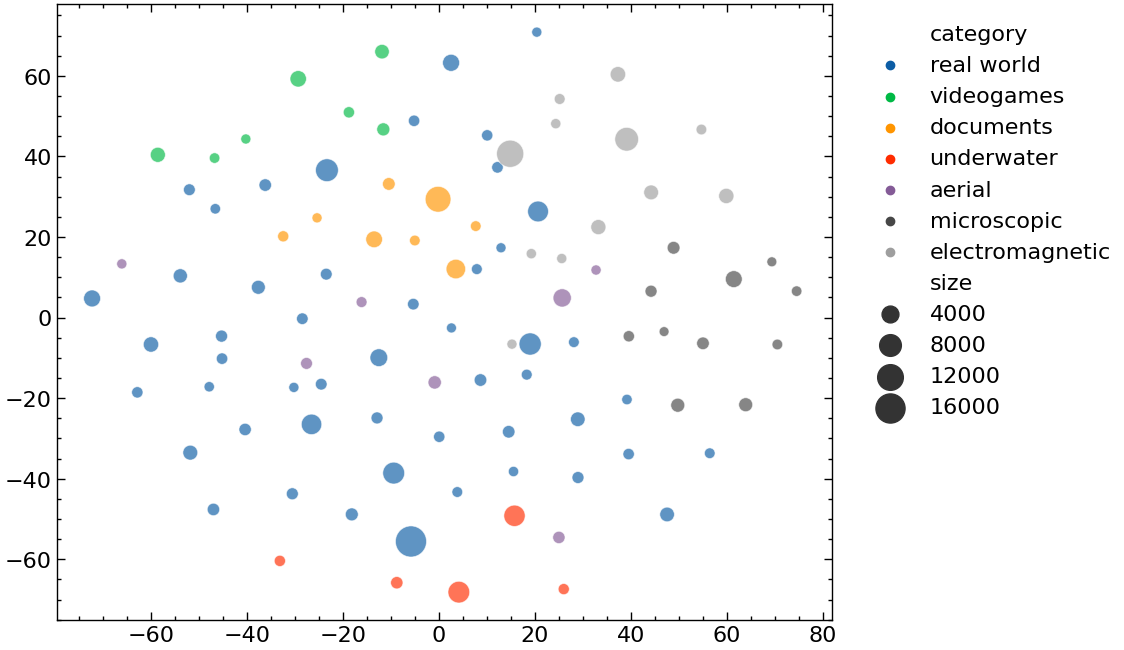}
    \caption{Scatter plot visualization of Roboflow-100 datasets CLIP vectors reduced to two dimensions via TSNE. The graph legend on the top right corner shows what category of dataset each color denotes. This visualization helps us determine the clustering degree of our collected datasets.}
    \label{fig:scatter-plot}
\end{figure}

%% file: tables/results-datasets-grouped.tex
\begin{table}[t]
    \begin{center}
    \setlength{\tabcolsep}{0.3em} % for the horizontal padding
    {\renewcommand{\arraystretch}{1.2}% for the vertical padding
        \begin{tabular}{ l | c c c}
            \hline
            Category        & YOLOv5       & YOLOv7       & GLIP         \\
            \hline
            Aerial          & 0.636        & 0.504        & 0.230         \\
            Videogames      & 0.859        & 0.796        & 0.188        \\
            Microscopic     & 0.650        & 0.591        & 0.159        \\
            Underwater      & 0.560        & 0.662        & 0.019        \\
            Documents       & 0.716        & 0.722        & 0.024        \\
            Electromagnetic & 0.689        & 0.607        & 0.058        \\
            Real World      & 0.752        & 0.699        & 0.108        \\
            \hline
            \textbf{Total}    & \textbf{0.694} & \textbf{0.654} & \textbf{0.112} \\
            \hline
        \end{tabular}
    }
    \caption{Experiments results on RF100. We recorded the average mAP@.50 value for the YOLOv5 and YOLOv7 models and the mAP@.50:.95 for the GLIP model for each category.}
    \label{table:results-datasets-grouped}
    \end{center}
\end{table}

%% file: figures/categories.tex
\begin{figure*}[h]
    \section{Appendix}
    \label{appendix:A}
    \begin{center}
    \subcaptionbox{\textbf{Areal} images gather from drone, space and static cameras \label{areal-sample}}
    {\includegraphics[width=1\linewidth]{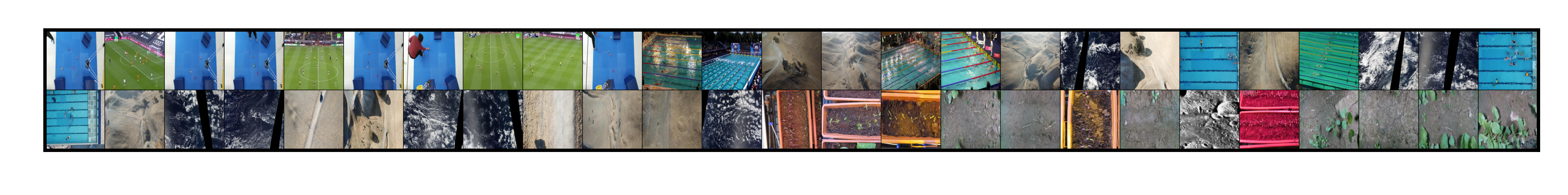}}
    \subcaptionbox{\textbf{Video Games} screen recordings from different games, such as Far Cry, Apex Legends, CS
        Go an League of Legends. \label{videogames-sample}}
    {\includegraphics[width=1\linewidth]{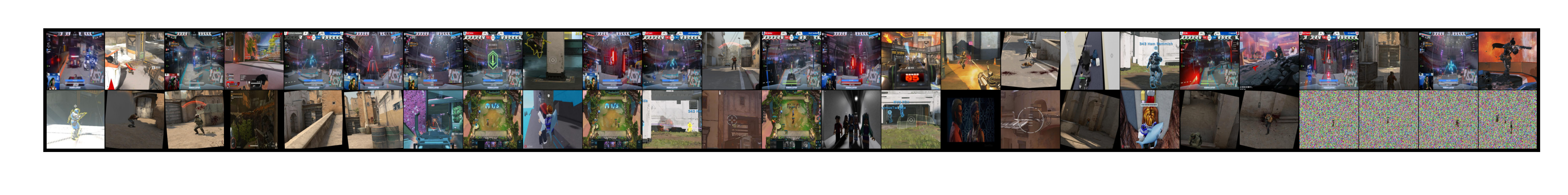}}
    \subcaptionbox{\textbf{Microscopic} images, mostly human diseases
        recorded with medical equipment showing cells, parasites, bacteria. \label{microscopic-sample}}
    {\includegraphics[width=1\linewidth]{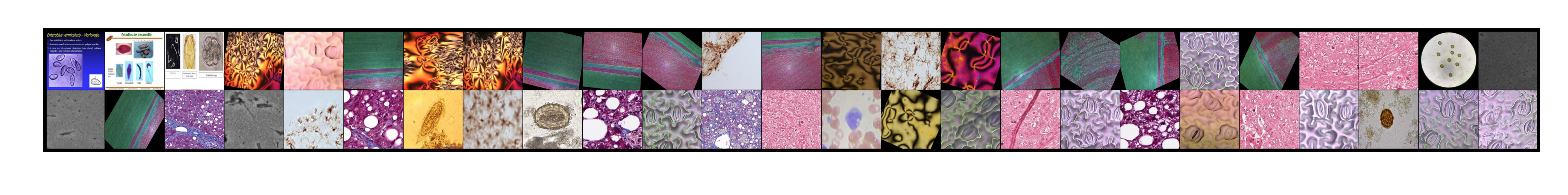}}
    \subcaptionbox{\textbf{Underwater} images of various sea plants
        and animal collected in the sea or aquariums. \label{underwater-sample}}
    {\includegraphics[width=1\linewidth]{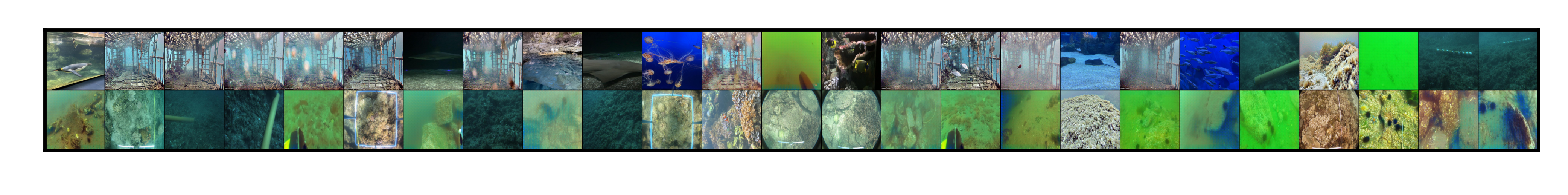}}
    \subcaptionbox{\textbf{Documents} such as tweets, tables and activity diagrams. \label{documents-sample}}
    {\includegraphics[width=1\linewidth]{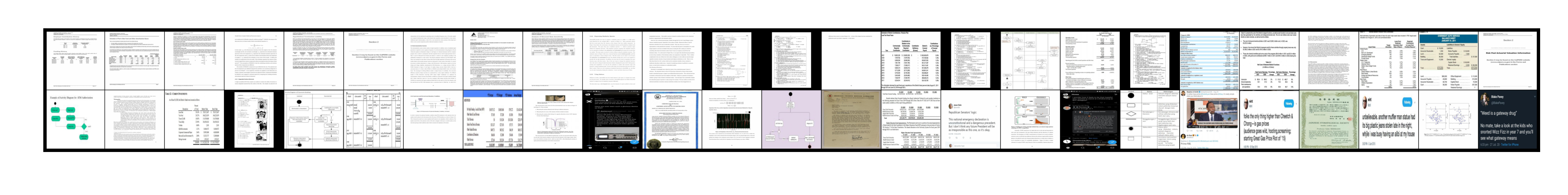}}
    \subcaptionbox{\textbf{Electromagnetic} images from X-ray/thermal cameras and MRI. \label{electromagnetic-sample}}
    {\includegraphics[width=1\linewidth]{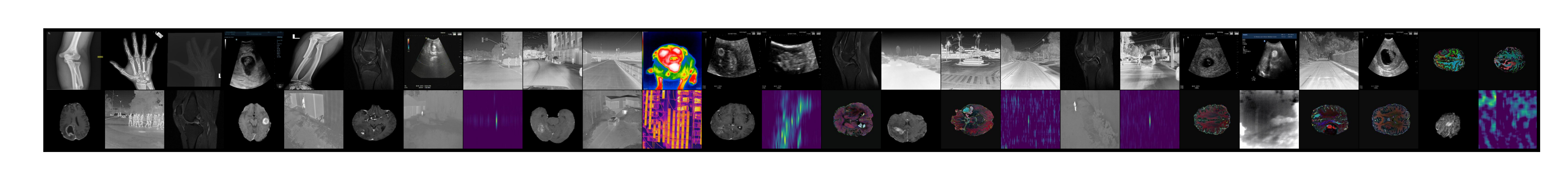}}
    \subcaptionbox{\textbf{Real World} images from a wide array of domains, animals, vehicles, human activities, paintings and electronics.
        \label{real-world-sample}}
    {\includegraphics[width=1\linewidth]{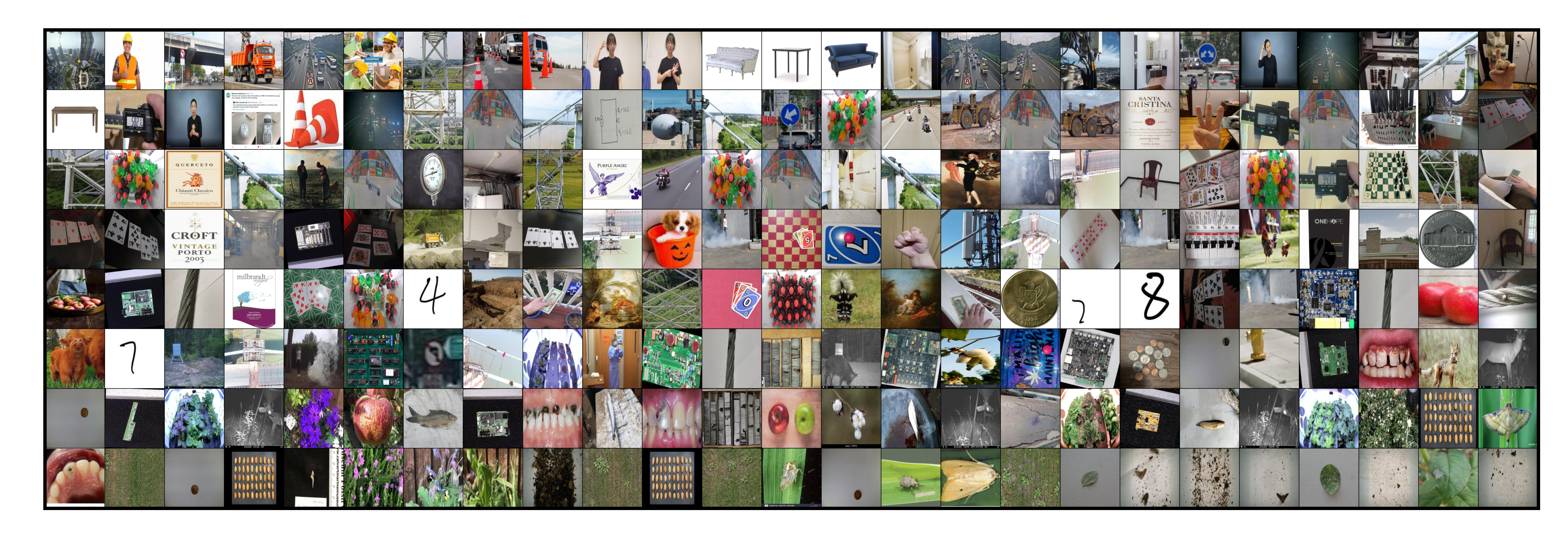}}
    \caption{Examples of images samples from different categories. \textit{Real World} was samples more due to its bigger size.}
    \label{fig:categories}
    \end{center}
\end{figure*}

%% file: tables/datasets-table.tex
%\begin{center}
\setlength{\tabcolsep}{0.3em} % for the horizontal padding
\LTcapwidth=\textwidth
\onecolumn
{\renewcommand{\arraystretch}{1.2}% for the vertical padding
    \begin{longtable}{| c | c | c c c | c c | c c c |}
        \hline
        Dataset                                                                          &
        Category                                                                         &    &
        Images                                                                                           &    &
        \multicolumn{2}{c}{Labeling}                                                                       &
        {YOLOv5}                                                                                           &
        {YOLOv7}                                                                                           &
        {GLIP}                \\                                                                            
           &    & Train & Valid & Test & Hours & Classes & mAP@.50 & mAP@.50 & mAP@.50:.95 
        \\ \hline
  \href{https://app.roboflow.com/roboflow-100/aerial-pool/3}{aerial pool}& \small{aerial} & 673 & 96 & 177 & 421 & 7 & 0.513 & \bfseries 0.791 & 0.013 \\

\href{https://app.roboflow.com/roboflow-100/secondary-chains/1}{secondary chains}& \small{aerial} & 103 & 16 & 43 & 201 & 1 & \bfseries 0.341 & 0.312 & 0.000 \\

\href{https://app.roboflow.com/roboflow-100/aerial-spheres/1}{aerial spheres}& \small{aerial} & 318 & 51 & 104 & 177 & 6 & \bfseries 0.993 & 0.539 & 0.000 \\

\href{https://app.roboflow.com/roboflow-100/soccer-players-5fuqs/1}{soccer players}& \small{aerial} & 114 & 16 & 33 & 0 & 3 & \bfseries 0.660 & 0.399 & 0.065 \\

\href{https://app.roboflow.com/roboflow-100/weed-crop-aerial/1}{weed crop}& \small{aerial} & 823 & 118 & 235 & 0 & 2 & \bfseries 0.820 & 0.615 & 0.027 \\

\href{https://app.roboflow.com/roboflow-100/aerial-cows/1}{aerial cows}& \small{aerial} & 1084 & 299 & 340 & 179 & 1 & \bfseries 0.854 & 0.568 & 0.056 \\

\href{https://app.roboflow.com/roboflow-100/cloud-types/1}{cloud types}& \small{aerial} & 3528 & 504 & 1008 & 0 & 4 & 0.271 & \bfseries 0.306 & 0.006 \\

\href{https://app.roboflow.com/roboflow-100/apex-videogame/1}{apex videogame} & \small{videogames} & 2583 & 415 & 691 & -1 & 2 & 0.839 & \bfseries 0.875 & n/a \\

\href{https://app.roboflow.com/roboflow-100/farcry6-videogame/1}{farcry6 videogame} & \small{videogames} & 82 & 14 & 24 & 0 & 11 & \bfseries 0.619 & 0.216 & 0.248 \\

\href{https://app.roboflow.com/roboflow-100/csgo-videogame/1}{csgo videogame} & \small{videogames} & 1774 & 207 & 446 & 0 & 2 & \bfseries 0.974 & 0.964 & 0.184 \\

\href{https://app.roboflow.com/roboflow-100/avatar-recognition-nuexe/1}{avatar recognition} & \small{videogames} & 225 & 30 & 59 & 3 & 1 & 0.889 & \bfseries 0.943 & 0.367 \\

\href{https://app.roboflow.com/roboflow-100/halo-infinite-angel-videogame/1}{halo infinite} & \small{videogames} & 462 & 71 & 136 & 4 & 4 & 0.921 & \bfseries 0.924 & 0.173 \\

\href{https://app.roboflow.com/roboflow-100/team-fight-tactics/1}{team fight} & \small{videogames} & 1162 & 112 & 307 & 88 & 59 & \bfseries 0.961 & 0.880 & 0.000 \\

\href{https://app.roboflow.com/roboflow-100/robomasters-285km/1}{robomasters 285km} & \small{videogames} & 1945 & 278 & 556 & 27 & 9 & \bfseries 0.816 & 0.772 & 0.003 \\

\href{https://app.roboflow.com/roboflow-100/stomata-cells/1}{stomata cells} & \small{microscopic} & 1482 & 209 & 414 & 0 & 2 & 0.840 & \bfseries 0.847 & 0.012 \\

\href{https://app.roboflow.com/roboflow-100/bccd-ouzjz/1}{bccd ouzjz} & \small{microscopic} & 255 & 36 & 73 & 0 & 3 & 0.912 & \bfseries 0.922 & 0.191 \\

\href{https://app.roboflow.com/roboflow-100/parasites-1s07h/1}{parasites 1s07h} & \small{microscopic} & 1484 & 215 & 411 & 0 & 8 & 0.848 & \bfseries 0.889 & 0.036 \\

\href{https://app.roboflow.com/roboflow-100/cells-uyemf/1}{cells uyemf} & \small{microscopic} & 16 & 2 & 4 & 210 & 1 & \bfseries 0.249 & 0.085 & 0.005 \\

\href{https://app.roboflow.com/roboflow-100/4-fold-defect/1}{4 fold} & \small{microscopic} & 503 & 33 & 134 & 279 & 1 & \bfseries 0.970 & 0.938 & 0.000 \\

\href{https://app.roboflow.com/roboflow-100/bacteria-ptywi/1}{bacteria ptywi} & \small{microscopic} & 30 & 10 & 10 & 472 & 1 & \bfseries 0.162 & 0.001 & 0.000 \\

\href{https://app.roboflow.com/roboflow-100/cotton-plant-disease/1}{cotton plant} & \small{microscopic} & 724 & 102 & 198 & 259 & 1 & \bfseries 0.204 & 0.052 & 0.000 \\

\href{https://app.roboflow.com/roboflow-100/mitosis-gjs3g/1}{mitosis gjs3g} & \small{microscopic} & 213 & 30 & 61 & 0 & 1 & \bfseries 0.931 & 0.739 & 0.001 \\

\href{https://app.roboflow.com/roboflow-100/phages/1}{phages} & \small{microscopic} & 1155 & 103 & 164 & 74 & 2 & \bfseries 0.854 & 0.842 & 0.002 \\

\href{https://app.roboflow.com/roboflow-100/liver-disease/1}{liver disease} & \small{microscopic} & 2782 & 400 & 794 & 31 & 4 & \bfseries 0.592 & 0.583 & 0.000 \\

\href{https://app.roboflow.com/roboflow-100/asbestos/1}{asbestos} & \small{microscopic} & 932 & 133 & 266 & 126 & 4 & 0.596 & \bfseries 0.611 & 0.007 \\

\href{https://app.roboflow.com/roboflow-100/underwater-pipes-4ng4t/1}{underwater pipes} & \small{underwater} & 5617 & 779 & 1575 & 316 & 1 & 0.995 & \bfseries 0.998 & 0.733 \\

\href{https://app.roboflow.com/roboflow-100/aquarium-qlnqy/1}{aquarium qlnqy} & \small{underwater} & 448 & 63 & 127 & 0 & 7 & 0.790 & \bfseries 0.822 & 0.195 \\

\href{https://app.roboflow.com/roboflow-100/peixos-fish/3}{peixos fish} & \small{underwater} & 821 & 118 & 261 & 0 & 12 & 0.148 & \bfseries 0.821 & 0.004 \\

\href{https://app.roboflow.com/roboflow-100/underwater-objects-5v7p8/1}{underwater objects} & \small{underwater} & 5320 & 760 & 1520 & 0 & 5 & \bfseries 0.693 & 0.453 & 0.005 \\

\href{https://app.roboflow.com/roboflow-100/coral-lwptl/1}{coral lwptl} & \small{underwater} & 427 & 74 & 93 & 165 & 14 & 0.174 & \bfseries 0.218 & 0.001 \\

\href{https://app.roboflow.com/roboflow-100/tweeter-posts/1}{tweeter posts} & \small{documents} & 87 & 9 & 21 & 0 & 2 & \bfseries 0.708 & 0.495 & 0.005 \\

\href{https://app.roboflow.com/roboflow-100/tweeter-profile/1}{tweeter profile} & \small{documents} & 425 & 61 & 121 & 0 & 1 & 0.988 & \bfseries 0.990 & 0.002 \\

\href{https://app.roboflow.com/roboflow-100/document-parts/1}{document parts} & \small{documents} & 906 & 150 & 318 & 192 & 2 & \bfseries 0.677 & 0.666 & 0.033 \\

\href{https://app.roboflow.com/roboflow-100/activity-diagrams-qdobr/1}{activity diagrams} & \small{documents} & 259 & 45 & 74 & 192 & 19 & 0.427 & \bfseries 0.509 & 0.002 \\

\href{https://app.roboflow.com/roboflow-100/signatures-xc8up/1}{signatures xc8up} & \small{documents} & 257 & 37 & 74 & 0 & 1 & \bfseries 0.961 & 0.932 & 0.082 \\

\href{https://app.roboflow.com/roboflow-100/paper-parts/3}{paper parts} & \small{documents} & 8472 & 1209 & 2359 & 211 & 46 & 0.590 & \bfseries 0.796 & 0.007 \\

\href{https://app.roboflow.com/roboflow-100/tabular-data-wf9uh/1}{tabular data} & \small{documents} & 3251 & 206 & 409 & 271 & 12 & 0.752 & \bfseries 0.782 & 0.018 \\

\href{https://app.roboflow.com/roboflow-100/paragraphs-co84b/1}{paragraphs co84b} & \small{documents} & 4209 & 633 & 1221 & 228 & 7 & \bfseries 0.626 & 0.610 & 0.000 \\

\href{https://app.roboflow.com/roboflow-100/thermal-dogs-and-people-x6ejw/1}{thermal dogs} & \small{electromagnetic} & 142 & 20 & 41 & -1 & 2 & \bfseries 0.967 & 0.957 & 0.470 \\

\href{https://app.roboflow.com/roboflow-100/solar-panels-taxvb/1}{solar panels} & \small{electromagnetic} & 112 & 19 & 30 & 175 & 5 & \bfseries 0.413 & 0.261 & 0.000 \\

\href{https://app.roboflow.com/roboflow-100/radio-signal/1}{radio signal} & \small{electromagnetic} & 1954 & 278 & 566 & 0 & 2 & \bfseries 0.673 & 0.653 & 0.066 \\

\href{https://app.roboflow.com/roboflow-100/thermal-cheetah-my4dp/1}{thermal cheetah} & \small{electromagnetic} & 90 & 14 & 25 & 0 & 2 & \bfseries 0.931 & 0.513 & 0.028 \\

\href{https://app.roboflow.com/roboflow-100/x-ray-rheumatology/1}{x ray} & \small{electromagnetic} & 135 & 16 & 34 & 16 & 12 & \bfseries 0.722 & 0.506 & 0.000 \\

\href{https://app.roboflow.com/roboflow-100/acl-x-ray/1}{acl x} & \small{electromagnetic} & 2141 & 306 & 612 & 0 & 1 & 0.995 & \bfseries 0.998 & 0.000 \\

\href{https://app.roboflow.com/roboflow-100/abdomen-mri/1}{abdomen mri} & \small{electromagnetic} & 1887 & 238 & 479 & 0 & 1 & \bfseries 0.965 & 0.958 & 0.021 \\

\href{https://app.roboflow.com/roboflow-100/axial-mri/1}{axial mri} & \small{electromagnetic} & 253 & 39 & 79 & 0 & 2 & \bfseries 0.638 & 0.549 & 0.039 \\

\href{https://app.roboflow.com/roboflow-100/gynecology-mri/1}{gynecology mri} & \small{electromagnetic} & 2122 & 253 & 526 & 7 & 3 & \bfseries 0.323 & 0.171 & 0.000 \\

\href{https://app.roboflow.com/roboflow-100/brain-tumor-m2pbp/1}{brain tumor} & \small{electromagnetic} & 6930 & 990 & 1980 & 0 & 3 & 0.768 & \bfseries 0.809 & 0.003 \\

\href{https://app.roboflow.com/roboflow-100/bone-fracture-7fylg/1}{bone fracture} & \small{electromagnetic} & 326 & 44 & 88 & 0 & 4 & 0.085 & \bfseries 0.090 & 0.000 \\

\href{https://app.roboflow.com/roboflow-100/flir-camera-objects/1}{flir camera} & \small{electromagnetic} & 9306 & 1452 & 2854 & 17 & 4 & 0.796 & \bfseries 0.824 & 0.073 \\

\href{https://app.roboflow.com/roboflow-100/hand-gestures-jps7z/1}{hand gestures} & \small{real world} & 642 & 94 & 178 & -1 & 14 & \bfseries 0.995 & 0.995 & n/a \\

\href{https://app.roboflow.com/roboflow-100/smoke-uvylj/1}{smoke uvylj} & \small{real world} & 522 & 76 & 148 & 7 & 1 & 0.959 & \bfseries 0.962 & 0.431 \\

\href{https://app.roboflow.com/roboflow-100/wall-damage/1}{wall damage} & \small{real world} & 325 & 40 & 96 & -1 & 3 & \bfseries 0.500 & 0.434 & n/a \\

\href{https://app.roboflow.com/roboflow-100/corrosion-bi3q3/1}{corrosion bi3q3} & \small{real world} & 840 & 105 & 304 & 186 & 3 & \bfseries 0.768 & 0.764 & 0.003 \\

\href{https://app.roboflow.com/roboflow-100/excavators-czvg9/1}{excavators czvg9} & \small{real world} & 2244 & 144 & 267 & 0 & 3 & \bfseries 0.946 & 0.895 & 0.274 \\

\href{https://app.roboflow.com/roboflow-100/chess-pieces-mjzgj/1}{chess pieces} & \small{real world} & 202 & 29 & 58 & 0 & 13 & \bfseries 0.977 & 0.830 & 0.017 \\

\href{https://app.roboflow.com/roboflow-100/road-signs-6ih4y/1}{road signs} & \small{real world} & 1376 & 229 & 488 & 0 & 21 & \bfseries 0.963 & 0.944 & 0.036 \\

\href{https://app.roboflow.com/roboflow-100/street-work/3}{street work} & \small{real world} & 611 & 87 & 175 & 2 & 11 & 0.478 & \bfseries 0.708 & 0.148 \\

\href{https://app.roboflow.com/roboflow-100/construction-safety-gsnvb/1}{construction safety} & \small{real world} & 997 & 90 & 119 & 505 & 5 & \bfseries 0.915 & 0.915 & 0.259 \\

\href{https://app.roboflow.com/roboflow-100/road-traffic/3}{road traffic} & \small{real world} & 494 & 133 & 187 & -1 & 12 & 0.597 & \bfseries 0.847 & n/a \\

\href{https://app.roboflow.com/roboflow-100/washroom-rf1fa/1}{washroom rf1fa} & \small{real world} & 1885 & 318 & 775 & 449 & 10 & 0.619 & \bfseries 0.634 & 0.146 \\

\href{https://app.roboflow.com/roboflow-100/circuit-elements/3}{circuit elements} & \small{real world} & 672 & 36 & 64 & 311 & 46 & \bfseries 0.063 & n/a & 0.001 \\

\href{https://app.roboflow.com/roboflow-100/mask-wearing-608pr/1}{mask wearing} & \small{real world} & 105 & 15 & 29 & 0 & 2 & \bfseries 0.788 & 0.513 & 0.008 \\

\href{https://app.roboflow.com/roboflow-100/cables-nl42k/1}{cables nl42k} & \small{real world} & 4816 & 794 & 1220 & 0 & 11 & 0.688 & \bfseries 0.722 & 0.010 \\

\href{https://app.roboflow.com/roboflow-100/soda-bottles/3}{soda bottles} & \small{real world} & 1547 & 216 & 486 & 243 & 6 & \bfseries 0.964 & n/a & 0.098 \\

\href{https://app.roboflow.com/roboflow-100/truck-movement/3}{truck movement} & \small{real world} & 740 & 107 & 215 & 282 & 7 & 0.786 & \bfseries 0.846 & 0.007 \\

\href{https://app.roboflow.com/roboflow-100/wine-labels/1}{wine labels} & \small{real world} & 3172 & 630 & 841 & 249 & 12 & 0.569 & \bfseries 0.632 & 0.045 \\

\href{https://app.roboflow.com/roboflow-100/digits-t2eg6/1}{digits t2eg6} & \small{real world} & 2912 & 367 & 824 & 144 & 10 & \bfseries 0.989 & 0.989 & 0.003 \\

\href{https://app.roboflow.com/roboflow-100/vehicles-q0x2v/1}{vehicles q0x2v} & \small{real world} & 2634 & 458 & 966 & 1121 & -1 & 0.454 & \bfseries 0.464 & 0.029 \\

\href{https://app.roboflow.com/roboflow-100/peanuts-sd4kf/1}{peanuts sd4kf} & \small{real world} & 268 & 42 & 77 & 212 & 2 & 0.995 & \bfseries 0.997 & 0.358 \\

\href{https://app.roboflow.com/roboflow-100/printed-circuit-board/3}{printed circuit} & \small{real world} & 548 & 44 & 80 & 311 & 34 & \bfseries 0.091 & n/a & 0.000 \\

\href{https://app.roboflow.com/roboflow-100/pests-2xlvx/1}{pests 2xlvx} & \small{real world} & 509 & 55 & 153 & 188 & 28 & \bfseries 0.136 & 0.029 & 0.004 \\

\href{https://app.roboflow.com/roboflow-100/cavity-rs0uf/1}{cavity rs0uf} & \small{real world} & 287 & 38 & 93 & 165 & 2 & 0.782 & \bfseries 0.799 & 0.029 \\

\href{https://app.roboflow.com/roboflow-100/leaf-disease-nsdsr/1}{leaf disease} & \small{real world} & 1589 & 296 & 616 & 143 & 3 & 0.531 & \bfseries 0.560 & 0.027 \\

\href{https://app.roboflow.com/roboflow-100/marbles/1}{marbles} & \small{real world} & 54 & 32 & 19 & 133 & 2 & \bfseries 0.992 & 0.473 & 0.030 \\

\href{https://app.roboflow.com/roboflow-100/pills-sxdht/1}{pills sxdht} & \small{real world} & 316 & 45 & 90 & 0 & 8 & \bfseries 0.869 & 0.867 & 0.194 \\

\href{https://app.roboflow.com/roboflow-100/poker-cards-cxcvz/1}{poker cards} & \small{real world} & 964 & 128 & 193 & 0 & 53 & \bfseries 0.886 & 0.251 & -0.000 \\

\href{https://app.roboflow.com/roboflow-100/number-ops/1}{number ops} & \small{real world} & 4869 & 623 & 1636 & 28 & 15 & 0.990 & \bfseries 0.992 & 0.055 \\

\href{https://app.roboflow.com/roboflow-100/insects-mytwu/1}{insects mytwu} & \small{real world} & 696 & 100 & 199 & 0 & 10 & \bfseries 0.890 & 0.858 & 0.024 \\

\href{https://app.roboflow.com/roboflow-100/cotton-20xz5/1}{cotton 20xz5} & \small{real world} & 367 & 20 & 19 & 17 & 4 & 0.569 & \bfseries 0.591 & 0.157 \\

\href{https://app.roboflow.com/roboflow-100/furniture-ngpea/1}{furniture ngpea} & \small{real world} & 454 & 74 & 161 & 0 & 3 & \bfseries 0.983 & 0.968 & 0.836 \\

\href{https://app.roboflow.com/roboflow-100/cable-damage/1}{cable damage} & \small{real world} & 919 & 134 & 265 & 2 & 2 & \bfseries 0.910 & 0.574 & 0.006 \\

\href{https://app.roboflow.com/roboflow-100/animals-ij5d2/1}{animals ij5d2} & \small{real world} & 700 & 100 & 200 & 12 & 10 & \bfseries 0.761 & 0.342 & 0.249 \\

\href{https://app.roboflow.com/roboflow-100/coins-1apki/1}{coins 1apki} & \small{real world} & 6121 & 699 & 1599 & 0 & 4 & 0.932 & \bfseries 0.977 & 0.175 \\

\href{https://app.roboflow.com/roboflow-100/apples-fvpl5/1}{apples fvpl5} & \small{real world} & 489 & 30 & 178 & -1 & 2 & 0.779 & \bfseries 0.791 & n/a \\

\href{https://app.roboflow.com/roboflow-100/people-in-paintings/1}{people in} & \small{real world} & 634 & 81 & 194 & 5 & 1 & 0.575 & \bfseries 0.678 & 0.168 \\

\href{https://app.roboflow.com/roboflow-100/circuit-voltages/1}{circuit voltages} & \small{real world} & 92 & 15 & 25 & 11 & 6 & \bfseries 0.797 & 0.257 & 0.009 \\

\href{https://app.roboflow.com/roboflow-100/uno-deck/1}{uno deck} & \small{real world} & 6295 & 899 & 1798 & 0 & 15 & 0.993 & \bfseries 0.994 & 0.013 \\

\href{https://app.roboflow.com/roboflow-100/grass-weeds/1}{grass weeds} & \small{real world} & 1661 & 245 & 580 & 105 & 1 & \bfseries 0.781 & 0.781 & 0.106 \\

\href{https://app.roboflow.com/roboflow-100/gauge-u2lwv/4}{gauge u2lwv} & \small{real world} & 158 & 25 & 52 & 141 & 2 & 0.642 & \bfseries 0.668 & 0.217 \\

\href{https://app.roboflow.com/roboflow-100/sign-language-sokdr/1}{sign language} & \small{real world} & 504 & 72 & 144 & 0 & 26 & \bfseries 0.870 & 0.255 & 0.006 \\

\href{https://app.roboflow.com/roboflow-100/valentines-chocolate/3}{valentines chocolate} & \small{real world} & 68 & 6 & 13 & 4 & 22 & \bfseries 0.110 & 0.059 & 0.013 \\

\href{https://app.roboflow.com/roboflow-100/fish-market-ggjso/5}{fish market} & \small{real world} & 14180 & 1202 & 3116 & 252 & 21 & 0.920 & \bfseries 0.988 & 0.013 \\

\href{https://app.roboflow.com/roboflow-100/lettuce-pallets/1}{lettuce pallets} & \small{real world} & 1060 & 151 & 299 & 168 & 5 & 0.945 & \bfseries 0.966 & 0.031 \\

\href{https://app.roboflow.com/roboflow-100/shark-teeth-5atku/1}{shark teeth} & \small{real world} & 191 & 36 & 53 & 154 & 4 & \bfseries 0.948 & 0.863 & 0.025 \\

\href{https://app.roboflow.com/roboflow-100/bees-jt5in/1}{bees jt5in} & \small{real world} & 5640 & 836 & 1604 & 163 & 1 & \bfseries 0.891 & 0.680 & 0.009 \\

\href{https://app.roboflow.com/roboflow-100/sedimentary-features-9eosf/4}{sedimentary features} & \small{real world} & 156 & 21 & 45 & 31 & 5 & \bfseries 0.327 & 0.244 & 0.000 \\

\href{https://app.roboflow.com/roboflow-100/currency-v4f8j/1}{currency v4f8j} & \small{real world} & 576 & 82 & 155 & 1 & 10 & \bfseries 0.583 & 0.514 & 0.099 \\

\href{https://app.roboflow.com/roboflow-100/trail-camera/1}{trail camera} & \small{real world} & 941 & 131 & 239 & 4 & 2 & 0.966 & \bfseries 0.969 & 0.512 \\

\href{https://app.roboflow.com/roboflow-100/cell-towers/1}{cell towers} & \small{real world} & 705 & 101 & 202 & 25 & 2 & 0.939 & \bfseries 0.942 & 0.053 \\

        \hline
        \caption{The above table reports metadata about each dataset in RF100. It includes each data-set's name, number of classes, labeling hours spend by the original author, the train/validation/test split used, model's mAP@.50 value on the three bench marked models and the source link. In the labeling hours column, a zero value denotes that the dataset was annotated outside of the Roboflow app, and a \textit{n/a} value denotes that the dataset was created before Roboflow started keeping track of the labeling hours data. 
            \label{table:datasets-stats} }
    \end{longtable}
}
%\end{center}

%% file: acl2020.bbl
\begin{thebibliography}{11}
\providecommand{\natexlab}[1]{#1}
\providecommand{\url}[1]{\texttt{#1}}
\expandafter\ifx\csname urlstyle\endcsname\relax
  \providecommand{\doi}[1]{doi: #1}\else
  \providecommand{\doi}{doi: \begingroup \urlstyle{rm}\Url}\fi

\bibitem[Deng et~al.(2009)Deng, Dong, Socher, Li, Li, and
  Fei-Fei]{deng2009imagenet}
Jia Deng, Wei Dong, Richard Socher, Li-Jia Li, Kai Li, and Li~Fei-Fei.
\newblock Imagenet: A large-scale hierarchical image database.
\newblock In \emph{2009 IEEE conference on computer vision and pattern
  recognition}, pages 248--255. Ieee, 2009.

\bibitem[Everingham et~al.(2010)Everingham, Gool, Williams, Winn, and
  Zisserman]{pascalVOC}
Mark Everingham, Luc~Van Gool, Christopher K.~I. Williams, John~M. Winn, and
  Andrew Zisserman.
\newblock The pascal visual object classes (voc) challenge.
\newblock \emph{Int. J. Comput. Vis.}, 88\penalty0 (2):\penalty0 303--338,
  2010.
\newblock URL
  \url{http://dblp.uni-trier.de/db/journals/ijcv/ijcv88.html#EveringhamGWWZ10}.

\bibitem[Jocher et~al.(2020)Jocher, Stoken, Borovec, NanoCode012,
  ChristopherSTAN, Changyu, Laughing, tkianai, Hogan, lorenzomammana, yxNONG,
  AlexWang1900, Diaconu, Marc, wanghaoyang0106, ml5ah, Doug, Ingham, Frederik,
  Guilhen, Hatovix, Poznanski, Fang, Yu, changyu98, Wang, Gupta, Akhtar,
  PetrDvoracek, and Rai]{glenn_jocher_2020_4154370}
Glenn Jocher, Alex Stoken, Jirka Borovec, NanoCode012, ChristopherSTAN, Liu
  Changyu, Laughing, tkianai, Adam Hogan, lorenzomammana, yxNONG, AlexWang1900,
  Laurentiu Diaconu, Marc, wanghaoyang0106, ml5ah, Doug, Francisco Ingham,
  Frederik, Guilhen, Hatovix, Jake Poznanski, Jiacong Fang, Lijun Yu,
  changyu98, Mingyu Wang, Naman Gupta, Osama Akhtar, PetrDvoracek, and Prashant
  Rai.
\newblock {ultralytics/yolov5: v3.1 - Bug Fixes and Performance Improvements},
  October 2020.
\newblock URL \url{https://doi.org/10.5281/zenodo.4154370}.

\bibitem[Kuznetsova et~al.(2020)Kuznetsova, Rom, Alldrin, Uijlings, Krasin,
  Pont-Tuset, Kamali, Popov, Malloci, Kolesnikov, Duerig, and
  Ferrari]{OpenImages}
Alina Kuznetsova, Hassan Rom, Neil Alldrin, Jasper Uijlings, Ivan Krasin, Jordi
  Pont-Tuset, Shahab Kamali, Stefan Popov, Matteo Malloci, Alexander
  Kolesnikov, Tom Duerig, and Vittorio Ferrari.
\newblock The open images dataset v4: Unified image classification, object
  detection, and visual relationship detection at scale.
\newblock \emph{IJCV}, 2020.

\bibitem[Li et~al.(2021)Li, Zhang, Zhang, Yang, Li, Zhong, Wang, Yuan, Zhang,
  Hwang, Chang, and Gao]{glip}
Liunian~Harold Li, Pengchuan Zhang, Haotian Zhang, Jianwei Yang, Chunyuan Li,
  Yiwu Zhong, Lijuan Wang, Lu~Yuan, Lei Zhang, Jenq-Neng Hwang, Kai-Wei Chang,
  and Jianfeng Gao.
\newblock Grounded language-image pre-training, 2021.
\newblock URL \url{https://arxiv.org/abs/2112.03857}.

\bibitem[Li et~al.(2022)Li, Zhang, Zhang, Yang, Li, Zhong, Wang, Yuan, Zhang,
  Hwang, et~al.]{li2022grounded}
Liunian~Harold Li, Pengchuan Zhang, Haotian Zhang, Jianwei Yang, Chunyuan Li,
  Yiwu Zhong, Lijuan Wang, Lu~Yuan, Lei Zhang, Jenq-Neng Hwang, et~al.
\newblock Grounded language-image pre-training.
\newblock In \emph{Proceedings of the IEEE/CVF Conference on Computer Vision
  and Pattern Recognition}, pages 10965--10975, 2022.

\bibitem[Lin et~al.(2014)Lin, Maire, Belongie, Bourdev, Girshick, Hays, Perona,
  Ramanan, Zitnick, and Dollár]{coco}
Tsung-Yi Lin, Michael Maire, Serge Belongie, Lubomir Bourdev, Ross Girshick,
  James Hays, Pietro Perona, Deva Ramanan, C.~Lawrence Zitnick, and Piotr
  Dollár.
\newblock Microsoft coco: Common objects in context, 2014.
\newblock URL \url{https://arxiv.org/abs/1405.0312}.

\bibitem[Radford et~al.(2021)Radford, Kim, Hallacy, Ramesh, Goh, Agarwal,
  Sastry, Askell, Mishkin, Clark, Krueger, and Sutskever]{clip}
Alec Radford, Jong~Wook Kim, Chris Hallacy, Aditya Ramesh, Gabriel Goh,
  Sandhini Agarwal, Girish Sastry, Amanda Askell, Pamela Mishkin, Jack Clark,
  Gretchen Krueger, and Ilya Sutskever.
\newblock Learning transferable visual models from natural language
  supervision, 2021.
\newblock URL \url{https://arxiv.org/abs/2103.00020}.

\bibitem[Shao et~al.(2019)Shao, Li, Zhang, Peng, Yu, Zhang, Li, and
  Sun]{objects365}
Shuai Shao, Zeming Li, Tianyuan Zhang, Chao Peng, Gang Yu, Xiangyu Zhang, Jing
  Li, and Jian Sun.
\newblock Objects365: A large-scale, high-quality dataset for object detection.
\newblock In \emph{2019 IEEE/CVF International Conference on Computer Vision
  (ICCV)}, pages 8429--8438, 2019.
\newblock \doi{10.1109/ICCV.2019.00852}.

\bibitem[Wang et~al.(2022)Wang, Bochkovskiy, and Liao]{yolov7}
Chien-Yao Wang, Alexey Bochkovskiy, and Hong-Yuan~Mark Liao.
\newblock Yolov7: Trainable bag-of-freebies sets new state-of-the-art for
  real-time object detectors, 2022.

\bibitem[Yuan et~al.(2021)Yuan, Chen, Chen, Codella, Dai, Gao, Hu, Huang, Li,
  Li, Liu, Liu, Liu, Lu, Shi, Wang, Wang, Xiao, Xiao, Yang, Zeng, Zhou, and
  Zhang]{florence}
Lu~Yuan, Dongdong Chen, Yi-Ling Chen, Noel Codella, Xiyang Dai, Jianfeng Gao,
  Houdong Hu, Xuedong Huang, Boxin Li, Chunyuan Li, Ce~Liu, Mengchen Liu,
  Zicheng Liu, Yumao Lu, Yu~Shi, Lijuan Wang, Jianfeng Wang, Bin Xiao, Zhen
  Xiao, Jianwei Yang, Michael Zeng, Luowei Zhou, and Pengchuan Zhang.
\newblock Florence: A new foundation model for computer vision, 2021.
\newblock URL \url{https://arxiv.org/abs/2111.11432}.

\end{thebibliography}
